\pdfoutput=1
\documentclass{fairmeta_seamless} %

\usepackage{amsmath,amsfonts,bm}

\def\eqref#1{equation~\ref{#1}}

\def\1{\bm{1}}

\DeclareMathAlphabet{\mathsfit}{\encodingdefault}{\sfdefault}{m}{sl}
\SetMathAlphabet{\mathsfit}{bold}{\encodingdefault}{\sfdefault}{bx}{n}

\usepackage{hyperref}
\usepackage{url}
\usepackage{graphicx}
\usepackage{enumitem}
\usepackage{algorithm,algorithmicx}
\usepackage[noend]{algpseudocode}
\algrenewcommand\algorithmicrequire{\textbf{Input:}}
\usepackage{booktabs}
\usepackage{multirow}
\usepackage[]{todonotes}
\usepackage{mathtools}
\usepackage{pgfplots}
\usepackage{listings}
\usepackage{pgfplots}
\usepackage{pgfplotstable}
\usepackage{soul}
\usepackage{subcaption}
\usepackage[utf8]{inputenc}
\usepackage[vietnamese, english]{babel}
\usepackage[font=small,labelfont=bf]{caption}
\usepackage[utf8]{inputenc} %
\usepackage[T1]{fontenc}    %
\usepackage{amsmath}
\usepackage{hyperref}       %
\usepackage{cleveref}
\usepackage{url}            %
\usepackage{booktabs}       %
\usepackage{amsfonts}       %
\usepackage{nicefrac}       %
\usepackage{inconsolata}    %
\usepackage{appendix}
\usepackage{etoolbox}
\usepackage{tablefootnote}
\usepackage{xspace}
\AtBeginEnvironment{appendices}{\crefalias{section}{appendix}}

\title{Efficient Monotonic Multihead Attention}

\author[]{Xutai Ma} 
\author[]{Anna Sun}
\author[\dagger]{Siqi Ouyang} 
\author[]{Hirofumi Inaguma} 
\author[]{Paden Tomasello}
\affiliation[]{FAIR at Meta}
\date{November 30, 2023}
\affiliation[\dagger]{UC Santa Barbara}
\correspondence{\email{xutaima@meta.com}}
\metadata[Code]{\url{https://github.com/facebookresearch/seamless_communication.}}

\newcommand{\simuleval}{\texttt{SimulEval}\xspace}
\newcommand{\writett}{\texttt{write}\xspace}
\newcommand{\readtt}{\texttt{read}\xspace}
\newcommand{\langcode}[1]{\texttt{#1}}
\newcommand{\mmafull}{Monotonic Multihead Attention}
\newcommand{\mma}{MMA}
\newcommand{\emmafull}{Efficient Monotonic Multihead Attention}
\newcommand{\emma}{EMMA}

\abstract{
We introduce the \emmafull{} (\emma{}), 
a state-of-the-art simultaneous translation model with numerically-stable and unbiased monotonic alignment estimation.
In addition, we present improved training and inference strategies,
including simultaneous fine-tuning from an offline translation model
and reduction of monotonic alignment variance.
The experimental results demonstrate that the proposed model attains state-of-the-art performance in simultaneous speech-to-text translation on the Spanish and English translation task. %
}
\begin{document}
\maketitle

\section{Introduction}

Simultaneous translation is a task focusing on reducing the latency of the machine translation system.
In this approach, a simultaneous translation model initiates the translation process even before the speaker completes their sentence.
This type of model plays a pivotal role in various low-latency scenarios, such as personal travels and international conferences,
where people desire seamless and real-time translation experiences.
In contrast to an offline model, 
which processes the entire input sentence and generates translation output in a single step, 
a simultaneous translation model operates on a partial input sequence.
The simultaneous model incorporates a policy mechanism to determine when the model should generate translation output. 
The policy is characterized by two actions: \readtt and \writett.
While the \writett action indicates that the model should generate a partial translation,
the \readtt action introduces a pause in the generation process,
allowing the model to acquire additional input information.
The simultaneous policy can be either rule-based or learned through training processes.

In recent times, 
within the domain of learned policies, 
a specific category known as monotonic attention-based policies \citep{raffel_online_2017, chiu_monotonic_2018, arivazhagan_monotonic_2019},
particularly Transformer-based \mmafull{} (\mma{}) \citep{ma_monotonic_2019}
has demonstrated state-of-the-art performance in simultaneous text-to-text translation tasks.
Monotonic attention provides an unsupervised policy learning framework that 
is based on an estimation of monotonic alignment during training time.
Despite \mma{}'s remarkable achievements in text-to-text translation, 
its adaptation to speech input faces certain challenges.
\cite{ma_simulmt_2020} revealed that the \mma{} model fails to yield significant improvements when applied to speech input over simple wait-k baseline.
\cite{ma_simulmt_2020} attributes the suboptimal performance of \mma{} on speech input to the granularity and continuity characteristics of the speech encoder states.

In this paper, we further investigate the adaptation of monotonic attention on speech translation.
We demonstrate the two primary factors underlying the sub-optimal performance. 
The first is the numerical instability and introduction of bias during monotonic alignment estimation, 
which comes from techniques introduced by \cite{raffel_online_2017}. 
The second is the significant variance in monotonic alignment estimation,
especially in the later part of the sentence,
due to the continuous nature of encoder states.
To address these challenges, we propose \emmafull{} (\emma{}). 
Specifically, the key contributions of this work include:
\begin{itemize}
	\item A novel numerically stable, unbiased monotonic alignment estimation, with yields state-of-the-art performance on both simultaneous text-to-text and speech-to-text translation.
	\item A new monotonic alignment shaping strategy, including new latency regularization and monotonic alignment variance reduction
	\item An enhanced training scheme, involving fine-tuning the simultaneous model based on a pre-trained offline model. 
\end{itemize}

\section{Background}
\subsection{Notation}
Given matrices $A$ and $B$, we annotate the operations used in later chapters, along with their implementation in PyTorch toolkit
in \Cref{tab:notations}.

\begin{table}[h]
    \centering
    \small
    \begin{tabular}{l|l|l}
    \toprule 
    Notation & Definition & PyTorch \\
    \midrule
    $A_{i,j}$ & Index $i$-th row and $j$-th column in matrix $A$ & \texttt{A[i, j]}\\
    $A_{i,:}$ & Index $i$-th row of $A$ as a vector& \texttt{A[[i], :]}  \\
    $A_{:,j}$ & Index $j$-th column of $A$ as a vector & \texttt{A[:, [j]]}\\
    $A \odot B$ & Element-wise product (Hadamard roduct) & \texttt{A * B} \\
    $A B$ & Matrix multiplication & \texttt{torch.bmm(A, B)} \\
    $\texttt{comprod}_l$(A) & Cumulative product on the $l$-th dimension & \texttt{torch.cumprod(A, dim=l)} \\
    $\texttt{comsum}_l$(A) & Cumulative summation on the $l$-th dimension & \texttt{torch.cumsum(A, dim=l)} \\
    $\texttt{triu}_b$(A) & Upper triangle of $A$ with a offset of $b$& \texttt{torch.triu(A, diagonal=b)} \\
    $J_{N\times M}$ & A matrix with size of $N$ by $M$, filled with 1 & \texttt{torch.ones(N, M)} \\
    $\texttt{roll}_k$ & Shift matrix by $k$ elements, on last dimension & \texttt{A.roll(k, dims=[-1])} \\
    \bottomrule
    \end{tabular}
    \caption{Matrix operations and their implementation in PyTorch}
    \label{tab:notations}
\end{table}

\subsection{Simultaneous Translation}
Denote $\mathbf{X}$ and $\hat{\mathbf{Y}}$ the input and output sequences of a translation system.
$\mathbf{X}$ is text token for text input and speech encoder states for speech input.
We introduce the concept of a delay sequence denoted as $\mathbf{D}$,
where each element $d_i$ is the length of input utilized in generating the corresponding output element $\hat{y}_i$.
It is essential to note that $\mathbf{D}$ forms a strictly monotonic non-decreasing sequence.

In a simultaneous translation system, there exists $\hat{\mathbf{Y}}$ such that $d_i < |\mathbf{X}|$.
Meanwhile, an offline translation means $d_i = |\mathbf{X}|$ for all $i$.
The measurement of $\mathbf{D}$ varies with input and output media.
In this paper, $d_i$ is measured in number of tokens for text input and seconds for speech input.

There are two aspects to evaluate simultaneous speech translation system: quality and latency.
While the quality evaluation is same as offline system,
for latency evaluation, we use the most commonly used metric Average Lagging (AL) \cite{ma_stacl_2019}, defined as
\begin{equation}
    \text{AL} = \frac{1}{\tau(|\mathbf{X}|)} \sum_{i=1}^{\tau(|\mathbf{X}|)} d_i - d^*_i    
    \label{chap:task-eq:al}
\end{equation}
where $\tau(|\mathbf{X}|) = \text{min}\{i | d_i=|\mathbf{X}|\}$ is the index of the first target translation when the policy first reaches the end of the source sentence. $d^*_i$ is the ideal policy defined as
\begin{equation}
	d^*_i = \left(i-1\right) \cdot \frac{|\mathbf{X}|}{|\mathbf{Y}|}
	\label{chap:task-eq:al-idea}
\end{equation}
where $\mathbf{Y}$ is the reference translation. 
As suggested by \citet{ma_simulmt_2020}, 
$|\mathbf{X}|$ is measured in number of source words for text input and in number of seconds of source speech for speech input.

\subsection{Monotonic Attention}
\label{sec:monotonic_attention}
Monotonic attention models
\citep{raffel_online_2017, chiu_monotonic_2018, arivazhagan_monotonic_2019, ma_monotonic_2019}
have a learnable policy based on monotonic alignment estimation during training time.
At a given time when $i-1$-th target translation has been predicted
and $j$-th source input has been processed,
a stepwise probability, denoted as $p_{i,j}$,
describes the likelihood the model is going to \writett the $i$-th prediction rather than \readtt the next input.
Specifically, it is defined as
 \begin{equation}
	p_{i, j} = P(\text{action} = \writett | i, j; \theta_p) = \texttt{Sigmoid}\left(\mathcal{N}_{\theta_p} \left(s_{i-1}, h_j\right)\right)
	\label{eq:p_choose}
\end{equation}
where $\mathcal{N}_{\theta_p}$ is the policy network, $s_{i-1}$ is $i-1$-th decoder state and $h_{j}$ is $j$-th encoder state.

\citet{raffel_online_2017} proposed an closed form estimation of alignments between source and target
$\alpha_{i,j}$ from $p_{i,j}$ during training:

\begin{equation}
    \alpha_{i,:} = p_{i,:} \odot \texttt{cumprod}_2(1 - p_{i,:}) \odot \texttt{cumsum}_2 \left(\alpha_{i-1,:} \odot \frac{1}{\texttt{cumprod}_2
    (1 - p_{i,:})}\right)
    \label{eq:alpha_rachel}
\end{equation}
While \cite{raffel_online_2017} handle the hard alignment between source and target,
\citet{chiu_monotonic_2018} introduce monotonic chunkwise attention (MoChA),
which enables soft attention in a chunk following the moving attention head.
\citet{arivazhagan_monotonic_2019} further proposed monotonic infinite lookback attention (MILk),
in which a soft attention is computed over all the previous history.
Given the energy $u_{i,j}$ for the $i$-th decoder state and the $j$-th encoder state,
an expected soft attention is calculated in \autoref{eq:milk_recurent}:
\begin{equation}
    \label{eq:milk_recurent}
    \beta_{i,j} = \sum_{k=j}^{|\mathbf{X}|} \left( \frac{\alpha_{i, k} \exp(u_{i,j})}{\sum_{l=1}^k  \exp(u_{i, l})} \right)
\end{equation}
where $\beta$ instead of $\alpha$ is then used in training.
\citet{arivazhagan_monotonic_2019} also introduce latency augmented training for latency control.
\citet{ma_monotonic_2019} further extend the monotonic attention to
multihead attention (\mma{}) for Transformer models.
The design of \mma{} is to enable every attention head as individual monotonic attention.

\section{\emmafull{}}
In this section, we will discuss three key factors of \emmafull{} (\emma{}):
numerically stable estimation, alignment shaping, and streaming finetuning.
It is noteworthy that the monotonic alignment estimation, denoted as $\alpha$, discussed in this section is based on a single attention head.
Following the same design as \cite{ma_monotonic_2019}, the same estimation of $\alpha$ is applied to every attention head in \mma{} as integrated into the Transformer \citep{vaswani_attention_2017} model.
Notably, only the infinite lookback \citep{arivazhagan_monotonic_2019} variant of monotonic attention is applied.
\subsection{Numerically Stable Estimation}

Similar to \cite{raffel_online_2017},
the objective of the monotonic estimation is to calculate the expected alignment $\alpha_{i, j}$,
from stepwise \writett action probability $p_{i, j}$.
Nevertheless, the numerical instability arises from the denominator in Equation \ref{eq:alpha_rachel}, 
particularly when dealing with the multiplication of several small probabilities.
To address this issue, we introduce a innovative numerically stable approach for estimation of monotonic attention.

In accordance with the approach proposed by \cite{raffel_online_2017}, 
the monotonic alignment between the $i$-th target item and the $j$-th source item can be represented as:  
\begin{equation}
    \alpha_{i,j} = p_{i,j} \sum_{k=1}^j \alpha_{i-1, k} \prod_{l=k}^{j-1}(1 - p_{i,l})
\end{equation}
This expression can be reformulated in a matrix multiplication format:
\begin{equation}
    \alpha_{i,:} = p_{i,:} \odot \alpha_{i-1, :} \mathbf{T}(i)  
    \label{eq:alpha_matrix_multi}
\end{equation}
where $\mathbf{T}(i)$ represents a transition matrix, with each of its elements defined as:
\begin{equation}
    \mathbf{T}(i)_{m,n} =
    \begin{cases}
    \prod_{l=m}^{n-1} (1 - p_{i,l}) & m < n\\
    1 & m = n\\
    0 & m > n \\
    \end{cases}
\end{equation}
$\mathbf{T}(i)_{m,n}$ indicates the probability of the policy at the $(i-1)$-th target step consecutively skipping from the $m$-th to the $n$-th inputs.
Moreover, the transition matrix can be further expressed as
\begin{equation}
	\mathbf{T}(i) = \texttt{triu}_0\left(\texttt{cumprod}_2(1 - \texttt{triu}_1\left( J_{|X| \times 1} \texttt{roll}_1(p_{i,:}) \right))\right)
\end{equation}
The operations within the equation can be efficiently executed in parallel on GPUs, 
as elaborated in \Cref{tab:notations}. 
Reframing Equation \ref{eq:alpha_matrix_multi}, we arrive at the following expression:
\begin{equation}
\alpha_{i,:} = p_{i,:} \odot \alpha_{i-1, :} \texttt{triu}_0\left(\texttt{cumprod}_2(1 - \texttt{triu}_1\left( J_{|X| \times 1} \texttt{roll}_1(p_{i,:}) \right))\right)
	\label{eq:emma}
\end{equation}
It is noteworthy that this estimation process is also closed-form, with the desirable properties of being numerically stable and unbiased, 
as it does not require a denominator as the product of probabilities within the equation.
A comprehensive derivation of this closed-form estimation is provided in \Cref{sec:appendix_estimation}.

\subsection{Alignment Shaping}
When training the infinite lookback variant of monotonic attention,
it is necessary to add latency regularization in order to prevent the model from learning a trivial policy.
Without latency regularization, 
the optimal policy to minimize cross-entropy loss is to \readtt the entire the sequence before starting the translation.
Therefore, we applied latency and variance regularizations to control the tradeoff between translation quality and latency of the learned simultaneous translation policy.

The latency of the alignment describes how much partial input information is needed by the model to generate part of the translation.
The reduction of latency is commonly achieved by introducing a regularization term derived from the estimated alignment.
Consistent with prior work, 
such as \cite{arivazhagan_monotonic_2019, ma_monotonic_2019},
the expected delays $\bar{\mathbf{D}}=\bar{d_1},...,\bar{d}_{|\mathbf{Y}|}$ are estimated from the expected alignment $\alpha$ during training time.
The expected delay of target token $y_i$, denoted as $\bar{d}_i$, is computed as
\begin{equation}
    \bar{d}_i = \texttt{E}[j | i] = \sum_{k=1}^{|\mathbf{X}|} k \alpha_{i,k}
\end{equation}

Given a latency metric $\mathcal{C}$, the loss term is then computed as
\begin{equation}
	\mathcal{L}_{\text{latency}} = \mathcal{C}(\bar{\mathbf{D}})
\end{equation}

The variance of the alignment characterizes the certainty of an estimation.
It is noteworthy that an alignment estimation can be low latency but high variance.
For instance, a random walk policy, which yields a monotonic alignment of linear latency,
has a huge variance on the estimation.
\cite{arivazhagan_monotonic_2019} proposed a method to reduce the uncertainty by introducing a Gaussian noise to the input of stepwise probability network.
Nevertheless, empirical results show that this method is not efficient, especially when applied to speech translation models.
Therefore, we propose an alternative regularization-based strategy.

Denote the $\mathcal{V} = \bar{v_1}, ..., \bar{v}_{|\mathbf{Y}|}$ as the expected variances of the monotonic alignment.
The expected variance of target token $y_i$, denoted as $\bar{v}_i$, can be expressed as
\begin{equation}
	\bar{v}_i = \texttt{E}[(j - \texttt{E}[j | i])^2 | i] = \texttt{E}[j^2 | i] -  \texttt{E}[j | i]^2 = \sum_{k=1}^{|\mathbf{X}|} k^2 \alpha_{i,k} - \left(\sum_{k=1}^{|\mathbf{X}|} k \alpha_{i,k}\right)^2
\end{equation}
We then introduce the alignment variance loss as the following:
\begin{equation}
	\mathcal{L}_{\text{variance}} = \sum_{i=1}^{|\mathbf{Y}|}  \bar{v}_i \
\end{equation}

To further reduce the alignment variance, we proposed an enhanced stepwise probability network as
\begin{equation}
	p_{i, j} = \texttt{Sigmoid}\left(\frac{ \texttt{FFN}_s(s_{i-1})^T \texttt{FFN}_h(h_j) + b }{\tau} \right)
\end{equation}
$\texttt{FFN}_s$ and $\texttt{FFN}_s$ serve as energy projections, constructed using multi-layer feedforward networks,
which increase the expressive capability of stepwise probability network over linear projection adopted by prior monotonic work.
$b$ is a learnable bias, initialized by a negative value.
 Its purpose is to schedule an easier the policy optimization process from the offline policy.
 $\tau$ is the temperature factor, to encourage polarized output from stepwise probability network.
 
 Finally, we optimize the model with the following objective
 \begin{equation}
 	\mathcal{L}(\theta) = -\text{log}(\mathbf{Y} | \mathbf{X}) + \lambda_{\text{latency}} \mathcal{L}_{\text{latency}} +  \lambda_{\text{variance}} \mathcal{L}_{\text{variance}}
 \end{equation}
 where $\lambda_{\text{latency}}$ and $\lambda_{\text{variance}}$ are the loss weights.

\subsection{Simultaneous Fine-tuning}
\label{sec:simul_finetuning}
In most prior work on simultaneous translation,
the model is usually trained from scratch.
However, this approach often requires substantial resources when dealing with extensive or multilingual scenarios.
For instance, it can be a significant challenge to retrain a simultaneous model with the configuration from recent large-scale multilingual models,
such as Whisper or SeamlessM4T. 
To leverage the recent advancements achieved with large foundational translation models and enhance the adaptability of the simultaneous translation model, we introduce a method for Simultaneous Fine-tuning.

Denote the an arbitrary offline encoder-decoder translation model as $\mathcal{M}(\theta^o_e, \theta^o_d)$, with $\theta_e$ representing the encoder parameters and $\theta_d$ representing the decoder parameters. 
The simultaneous model is denoted as $\mathcal{M}(\theta_e, \theta_d, \theta_p)$, where $\theta_p$ denotes the policy network.
Simultaneous fine-tuning involves initializing $\theta_e$ with $\theta^o_e$ and $\theta_d$ with $\theta^o_d$. 
During the training process, the encoder parameters $\theta_e$ remain fixed, and optimization is only performed on $\theta_d$ and $\theta_p$. 
This design is motivated by the assumption that the generative components of the model, 
namely $\theta_e$ and $\theta_d$, 
should closely resemble those of the offline model. 
In simultaneous setting, they are adapted to partial contextual information.

\subsubsection{Streaming Inference}
\label{sec:streaming.inference}
We used \simuleval~\citep{ma_simuleval_2020} to build the inference pipeine.
The overall inference algorithm is illustrated in \Cref{algo:streaming.inference}.
For streaming speech input, we update the whole encoder every time a new speech chunk is received by the model.
Then, we run the decoder to generate a partial text translation based on the policy.

\begin{algorithm}
  \caption{\emma{} Inference}
  \label{algo:streaming.inference}
  \begin{algorithmic}[1]
    \Require{$\mathbf{X}$: Input streaming speech.}
    \Require{$\mathbf{Y}$: Output text.}
    \Require{$t_{\text{\emma{}}}$ : Decision threshold for \emma{}}
    \State $i \gets 1$, $j \gets 0$, $k \gets 0$
    \State $s_0 \gets \texttt{TextDecoder}(y_{0})$
    \While{$y_{i-1} \neq \texttt{EndOfSequence}$}
        \State $j \gets j + 1$
        \State $h_{\leq j} \gets \texttt{SpeechEncoder}(\mathbf{X}_{\leq j})$
        \While{$y_{i-1} \neq \texttt{EndOfSequence}$}
            \State $p \gets 1$
            \For{$\texttt{StepwiseProbabilty}$ in all attention head}
            \State $p \gets \texttt{min}(p, \texttt{StepwiseProbabilty}(h_j, s_{i-1}))$
            \EndFor
            \If{$p < t_{\text{\emma{}}}$}
            \State Break
            \Else
                \State $y_i, s_i \gets \texttt{TextDecoder}(s_{<i}, h_{\leq j})$
                \State $k \gets k + 1$
                \State $i \gets i + 1$
            \EndIf
            \EndWhile
    \EndWhile
  \end{algorithmic}
\end{algorithm}

\section{Experimental Setup}
We evaluate proposed models on speech-to-text translation task.
The models are evalauted with the \simuleval~\citep{ma_simuleval_2020} toolkit.
Evaluation of the models focuses on two factors: quality and latency. 
The quality is measured by detokenized BLEU,
using the SacreBLEU \citep{post_call_2018} toolkit.
Latency evaluation is measured by Average Lagging (AL) \citep{ma_stacl_2019}
We follow the simultaneous fine-tuning strategy introduced in \Cref{sec:simul_finetuning}.
The simultaneous model is initialized from an offline translation model.
Detailed information regarding the tasks, 
evaluation datasets employed in this study, 
and the performance of the offline model are presented in \Cref{tab:offline}

\begin{table}[h]
    \centering
    \small
    \begin{tabular}{l|l|l|l}
    \toprule 
    Task & Evaluation Set & Language & BLEU  \\
    \midrule
    \multirow{2}{*}{Bilingual} & mTedX & \langcode{spa-eng} & 37.1  \\
     & Must-C & \langcode{eng-spa}  & 38.1\\
    \midrule
    Multilingual & Fleurs & \langcode{100-eng} & 28.8 \tablefootnote{Average on 100 language directions} \\
    \bottomrule
    \end{tabular}
    \caption{Offline model performance}
    \label{tab:offline}
\end{table}

For speech-to-text (S2T) translation task,
we establish two experimental configurations: bilingual and multilingual.

The bilingual setup aims to demonstrate the model's potential when provided with a extensive corpus of training data. 
We trained one model for each direction, \langcode{spa-eng} and \langcode{eng-spa}.
The multilingual task demonstrates the model's capacity for rapid adaptation in offline-to-simultaneous transition, from an existing large scale multilingual translation model, SeamlessM4T \citep{seamlessm4t2023}.

In the bilingual setting, we follow the data setting from \cite{inaguma-etal-2023-unity}.
In the multilingual setting, we use the speech-to-text data from labeled and pseudo-labeled data in \cite{seamlessm4t2023}

In the bilingual S2T setup, we initialize the offline model with a pre-trained wav2vec 2.0 encoder \citep{baevski_wav2vec_2020} and mBART  decoder \citep{liu_multilingual_2020}. 
Subsequently, we initialize the simultaneous model based on this pre-trained offline model. 
The bilingual model is trained on supervised on semi-unsupervised data.
In the multilingual setting, we initialize the simultaneous model with the S2T part of an offline SeamlessM4T model, trained with the same labeled and pseudo-labeled data,
and evaluate the model on \langcode{100-eng} directions.

\section{Related Work}
Recent research has focused on the neural end-to-end approach, anticipating that a simpler system can reduce errors between subsystems and enhance overall efficiency in direct translation. Initially applied to text translation, this approach extended to speech-to-text tasks, showing competitiveness against cascade approaches. \citet{duong_attentional_2016} introduced an attention-based sequence-to-sequence structure for speech-to-text, using a recurrent neural network (RNN) based encoder-decoder architecture. Despite the novelty, there was a significant quality downgrade compared to cascaded approaches. Subsequent studies \cite{berard_listen_2016,weiss_sequence--sequence_2017,bansal_low-resource_2018,berard_end--end_2018} added convolutional layers, significantly improving end-to-end model performance. Leveraging the success of the Transformer in text translation \cite{vaswani_attention_2017}, \citet{di_gangi_enhancing_2019} and \citet{inaguma_espnet-st_2020} applied it to speech translation, achieving further improvements in quality and training speed.

Simultaneous translation policies are categorized into three groups. The first category consists of predefined context-free rule-based policies. \citet{cho_can_2016} proposed a Wait-If-* policy for offline simultaneous decoding, later modified by \citet{dalvi_incremental_2018} for consecutive prediction. Another variation, the Wait-$k$ policy, was introduced by \citet{ma_stacl_2019}, where the model alternates between reading $k$ inputs and performing read-write operations. The second category involves a learnable flexible policy with an agent, applying reinforcement learning. Examples include \citet{grissom_ii_dont_2014}, who used a Markov chain-based agent for phrase-based machine translation, and \citet{gu_learning_2017}, who introduced an agent that learns translation decisions from interaction with a pre-trained neural machine translation model. The third category features models using monotonic attention, replacing Softmax attention and leveraging closed-form expected attention. Notable works include \citet{raffel_online_2017}, \citet{chiu_monotonic_2018}, \citet{arivazhagan_monotonic_2019}, and \citet{ma_monotonic_2019}, demonstrating advancements in online linear time decoding and translation quality improvements.

\section{Results}
\subsection{Quality-Latency Trade-off}
\begin{figure}[h]
     \centering
     \begin{subfigure}[b]{0.4\textwidth}
         \centering
         \includegraphics[width=\textwidth]{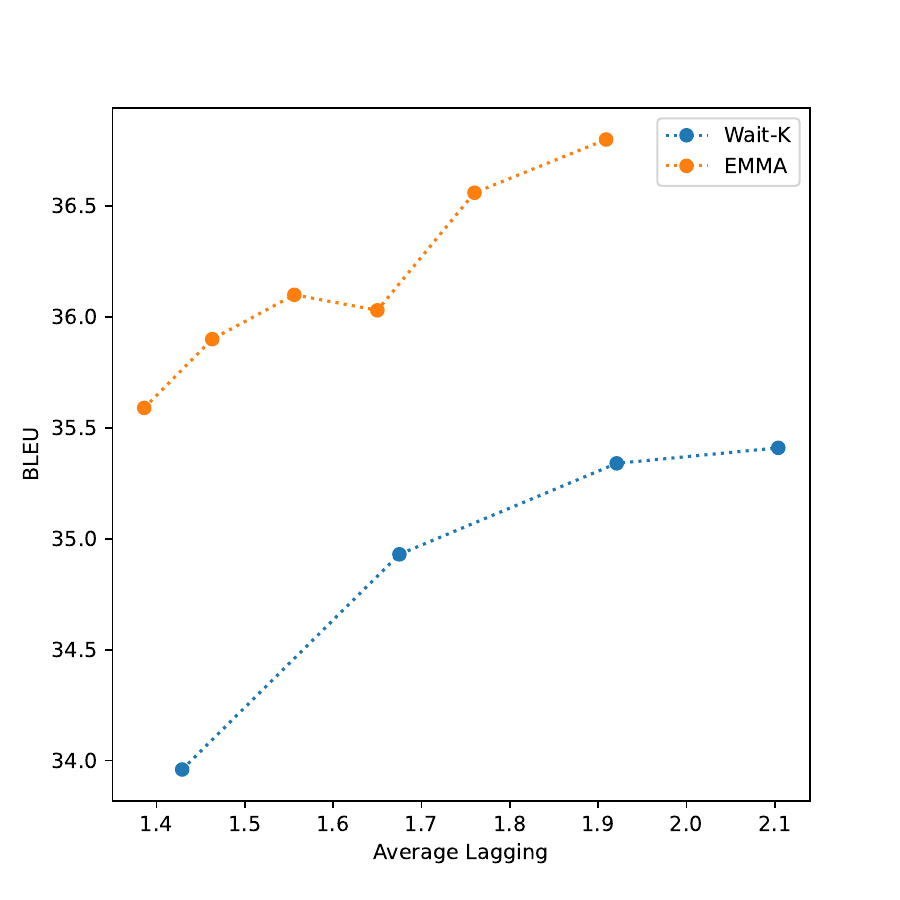}
         \caption{Spanish to English}
         \label{fig:es-en}
     \end{subfigure}
     \begin{subfigure}[b]{0.4\textwidth}
         \centering
         \includegraphics[width=\textwidth]{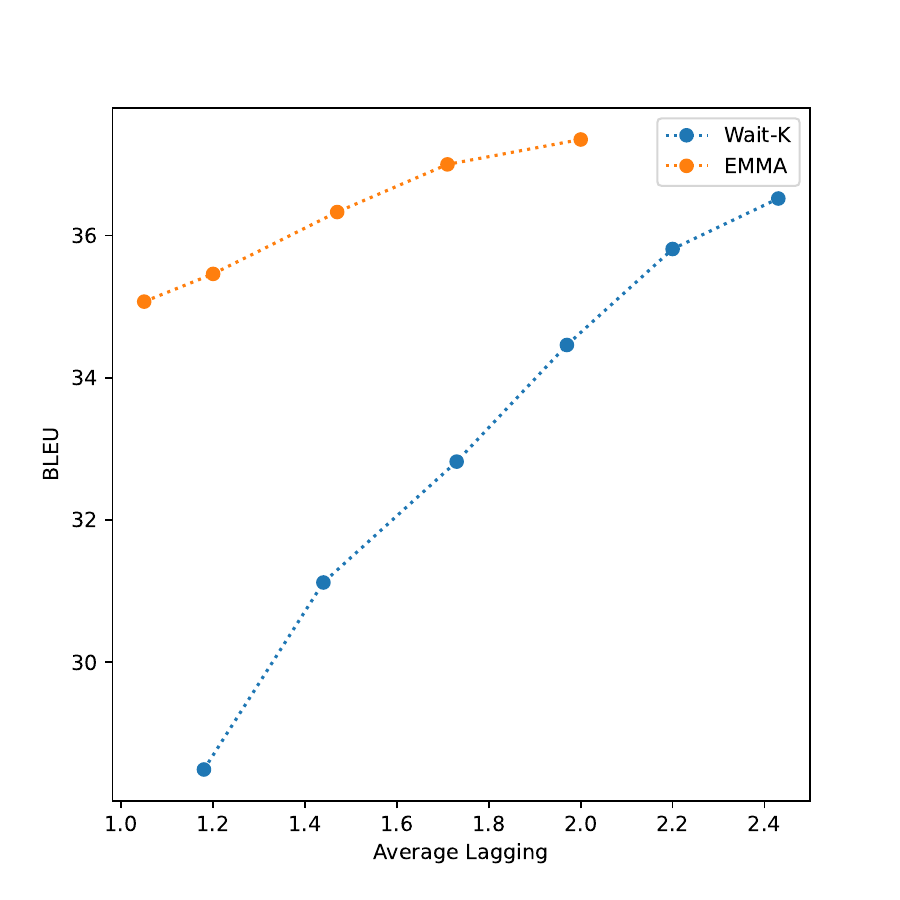}
         \caption{English to Spanish}
         \label{fig:en-es}
     \end{subfigure}
      \caption{The quality-latency trade-off for bilingual model.}
      \label{fig:quality-latency}
\end{figure}

We present the quality-latency trade-off on the bilingual setting.
\Cref{fig:quality-latency} shows the BLEU score under different latency settings.
We can see that the \emma{} model significantly outperforms the Wait-k model on all the latency regions in both directions.

We present the quality-latency trade-off on the multilingual setting, in \Cref{tbl:mtl.latency-quality}.
SeamlessM4T-\emma{} can achieve decent translation quality in a much shorter training time compared with training from scratch.
\begin{table}[!tbh]
    \centering
    \small
    \begin{tabular}{@{}lccc@{}}
        \toprule
         &  $t_{\text{\emma{}}}$ & BLEU & AL  \\
        \midrule
        SeamlessM4T&  & 28.8 & - \\\midrule
        \multirow{4}{*}{SeamlessM4T-\emma{}}  
        & 0.5 & 26.0 & 1.75 \\
        & 0.7 & 26.4 & 1.88 \\
        & 0.4 & 25.9 & 1.68 \\
        & 0.6 & 26.2 & 1.81 \\
        \bottomrule
    \end{tabular}
    \caption{Average translation quality and latency under multilingual setting with different latency decision thresholds $t_{\text{\emma{}}}$.}
    \label{tbl:mtl.latency-quality}
\end{table}

\begin{figure}[!tbh]
     \centering
         \includegraphics[width=\textwidth]{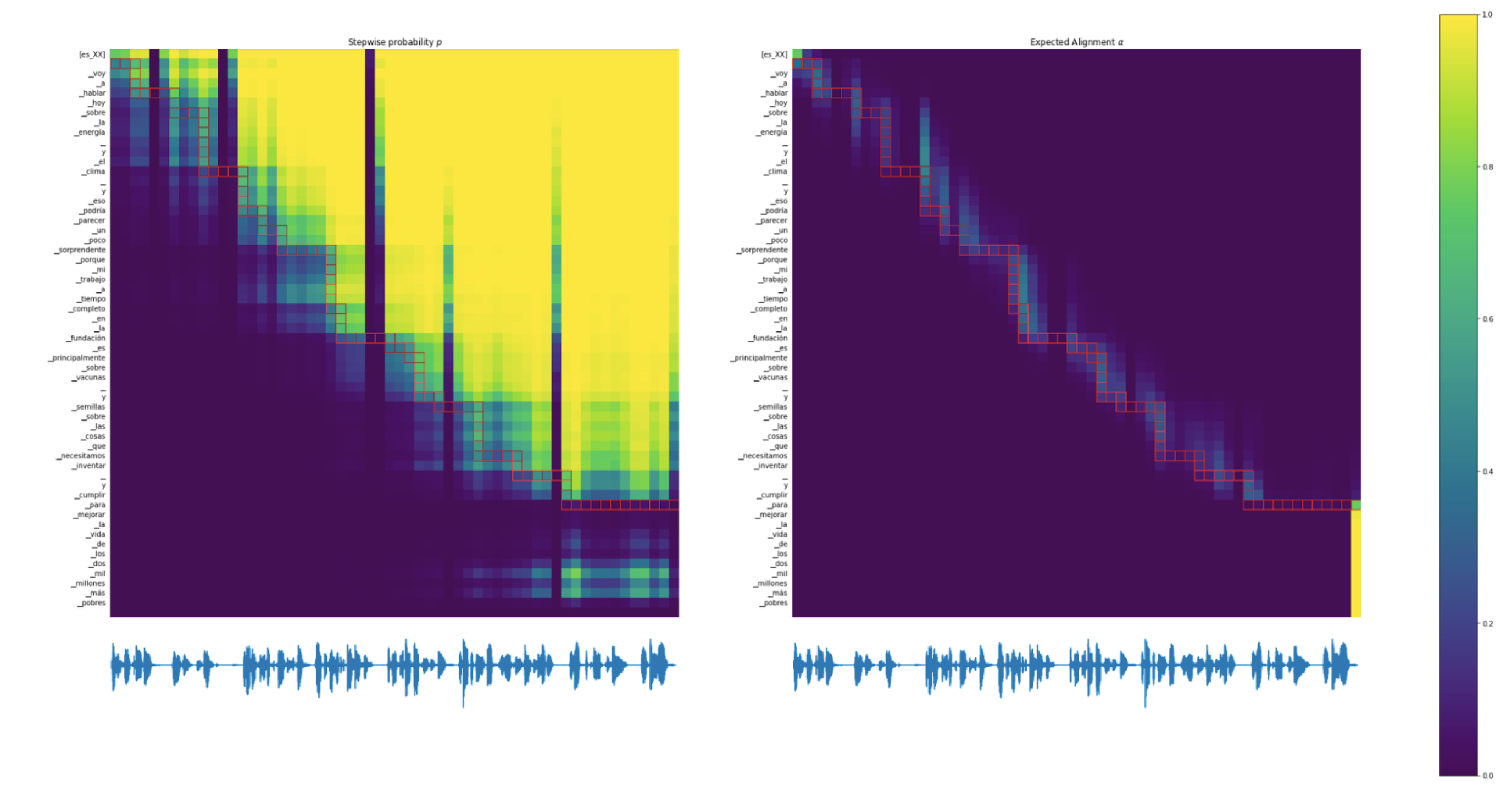}
         \caption{Visualization of monotonic alignment. Left side is stepwise probability $p$ and right side is estimated alignment $\alpha$.}
         \label{fig:policy}
\end{figure}
\subsection{Visualization}
We also visualize the policy learned from \emma{} as \Cref{fig:policy}. 
We can see that with the guidance of latency and variance regularization, the model can learn a monotonic policy aligning input speech and output text.

\section{Conclusion}
We propose \emmafull{} (\emma{}), with a novel numerically stable alignment estimation.
We fine-tune our model initialized from an offline model to speed up training and improve the performance.
We evaluate our model in bilingual and multilingual settings, and observe improvement over the baseline in both.

\bibliography{reference}
\bibliographystyle{iclr2023_conference}

\newpage
\appendix
\section*{Appendix}
\begin{appendices}
\section{Numerically Stable Estimation}
\label{sec:appendix_estimation}
Intuitively the $\alpha$ can be estimated from dynamic programming:
\begin{equation}
    \alpha_{i,j} = p_{i,j} \sum_{k=1}^j \alpha_{i-1, k} \prod_{l=k}^{j-1}(1 - p_{i,l})
    \label{eq:alpha_orig}
\end{equation}

While \Cref{eq:alpha_rachel} gives an closed form and parallel estimation of alignment,
the denominator in the equation can cause instability and alignment vanishing in the training.
We rewrite \Cref{eq:alpha_orig} as 
\begin{equation}
    \alpha_{i,:} = p_{i,:} \odot \alpha_{i-1, :} \mathbf{T}(i)  
\end{equation}
where $\mathbf{T}(i)$ a transition matrix and each of its elements are defined as:
\begin{equation}
    \mathbf{T}(i)_{m,n} =
    \begin{cases}
    \prod_{l=m}^{n-1} (1 - p_{i,l}) & m < n\\
    1 & m = n\\
    0 & m > n \\
    \end{cases}
\end{equation}
$\mathbf{T}(i)_{m,n}$ is the probability of a \readtt from $x_m$ to $x_n$ with $y_{i}$ without \writett.
Denote $t^{i}_{m,n} = \prod_{l=m}^{n} (1 - p_{i,l}) $
We can see that if we manage to have $\mathbf{T}(i)$, then the
$\alpha_{i, :}$ can be simply computed through matrix multiplication.

Define the probability of jumping from $x_m$ to $x_n$ with our write a new token $y_i$:

then we can expand $\mathbf{T}(i)$ as
\begin{equation}
    \mathbf{T}(i) = \begin{pmatrix}
1 & t^{i}_{1,2} & t^{i}_{1,3} & t^{i}_{1,4}  &... & & t^{i}_{1,|X|}\\\
0 & 1 & t^{i}_{2,3} & t^{i}_{2,4} &... & & t^{i}_{2,|X|}\\
0 & 0 & 1 & t^{i}_{3,4} &... & & t^{i}_{3,|X|}\\
\vdots & \vdots & \vdots & \vdots  & & & \vdots \\
0 & 0 & 0 & 0 &... & & 1\\
\end{pmatrix}_{|X| \times |X|}
\end{equation}
It can be further expressed as
\begin{align}
    \mathbf{T}(i) &=
     \texttt{triu}_0 \left(\begin{pmatrix}
1 & t^{i}_{1,2} & t^{i}_{1,3} & t^{i}_{1,4}  &... & & t^{i}_{1,|X|}\\\
1 & 1 & t^{i}_{2,3} & t^{i}_{2,4} &... & & t^{i}_{2,|X|}\\
1 & 1 & 1 & t^{i}_{3,4} &... & & t^{i}_{3,|X|}\\
\vdots & \vdots & \vdots & \vdots  & & & \vdots \\
1 & 1 & 1 & 1 &... & & 1\\
\end{pmatrix}_{|X| \times |X|} \right)\\
&= \texttt{triu}_0\left(\texttt{cumprod}_2(1 - \mathbf{P}^{ext}(i))\right)
\end{align}
where $\texttt{triu}_b\left(\cdot\right)$
is function to extract the upper triangle of a matrix with an offset $b$ \footnote{See $\texttt{torch.triu}$},
and $\texttt{cumprod}_2$ means that the computation is along the second dimension.
Additionally, the extended probability matrix $\mathbf{P}_i^{ext}$ is defined as
\begin{align}
    \mathbf{P}^{ext}(i)  &= \begin{pmatrix}
0 & p_{i,1} & p_{i,2}& ... & p_{i,|X| - 1}\\
0 & 0 & p_{i,2}& ... & p_{i,|X| - 1}\\
0 & 0 & 0& ... & p_{i,|X| - 1}\\
\vdots & \vdots & \vdots & & \vdots\\
0 & 0& 0& ... & p_{i,|X| - 1}\\
0 & 0& 0& ... & 0\\
\end{pmatrix}_{|X| \times |X|}\\
    &= \texttt{triu}_1 \left( \begin{pmatrix} 1 \\ \vdots \\ 1 \end{pmatrix}_{|X| \times 1}  \begin{pmatrix} p_{i,|X|} &p_{i,1} & ... & p_{i,|X| - 1} \end{pmatrix}_{1 \times |X|}\right)\\
    &= \texttt{triu}_1\left( J_{|X| \times 1} \texttt{roll}_1(p_{i,:})\right)
\end{align}
Where $J_{|X| \times 1}$ is an all one matrix with a size of $|X|$ by 1,
\footnote{$J_{|X| \times 1} \texttt{roll}_1(p_{i,:})$ can be achived by $\texttt{torch.expand}$ function.}
and $\texttt{roll}_k$ is the function to shift matrix by $k$ elements \footnote{See \texttt{torch.roll}}.

In summary, we can rewrite \Cref{eq:alpha_orig} as
\begin{equation}
    \alpha_{i,:} = p_{i,:} \odot \alpha_{i, :} \texttt{triu}_0\left(\texttt{cumprod}_2(1 - \texttt{triu}_1\left( J_{|X| \times 1} \texttt{roll}_1(p_{i,:}) \right))\right)
\end{equation}

A code snippet of implementation of \emma{} in PyTorch is shown as follows:
\begin{verbatim}

def monotonic_alignment(p):
    bsz, tgt_len, src_len = p.size()

    # Extension probablity matrix
    p_ext = p.roll(1, [-1]).unsqueeze(-2).expand(-1, -1, src_len, -1).triu(1)

    # Transition matrix
    T = (1 - p_ext).comprod(-1).triu()

    alpha = [p[:, [0]] * T[:, [0]]

    for i in range(1, tgt_len):
        alpha.append(p[:, [i]] * torch.bmm(alpha[i - 1], T[:, i]))
        
    return torch.cat(alpha[1:], dim=1)


\end{verbatim}
\end{appendices}

\end{document}